\documentclass[11pt]{article}

\usepackage[preprint]{acl}
\usepackage{times}
\usepackage{latexsym}

\usepackage[T1]{fontenc}

\usepackage[utf8]{inputenc}

\usepackage{microtype}

\usepackage{inconsolata}

\usepackage{graphicx}

\usepackage{tabularx}
\usepackage{booktabs}
\usepackage{longtable}
\usepackage{multirow}
\usepackage{xcolor}
\usepackage{subcaption}
\usepackage{float}
\usepackage{caption}

\title{GRADE: Generalizable Reasoning-Aware Dialogue Evaluation for AI Tutors}

\author{
  \textbf{Parth Bhalerao}\thanks{Equal contribution.} \quad
  \textbf{Jeromy Chang}\footnotemark[1] \quad
  \textbf{David Chou}\footnotemark[1] \quad
  \textbf{Oana Ignat} \\
  Santa Clara University \\
  \texttt{\{pbhalerao, jchang5, dchou, oignat\}@scu.edu}
}

\begin{document}
\maketitle
\begin{abstract}
Evaluating AI tutor responses requires more than factual correctness: tutors must identify mistakes, locate errors, provide guidance, and offer 
actionable next steps. We present GRADE, a systematic study of open-source 
models for pedagogical ability assessment in student-tutor dialogues. 
Building on the BEA 2025 TutorMind setting, we evaluate 120 configurations 
across five language models, zero-shot inference, LoRA fine-tuning, synthetic 
augmentation, CoT+Reasoning, and single-task versus multitask formulations. 
Gemma3-12B performs best for single-task evaluation, while Gemma3-27B in 
8-bit precision is more reliable for multitask prediction. We find that 
augmentation helps models that struggle with the original data, verification 
adds limited gains despite higher cost, and CoT+Reasoning is more useful 
for synthetic data generation than direct classification. We further show 
that LoRA fine-tuning on structured classification objectives interferes 
with instruction-following behavior under thinking mode, redirecting 
generation away from the required evaluation format. Carbon analysis shows 
that model choice and reasoning mode substantially affect emissions. Overall, 
GRADE shows that carefully selected open-source LoRA pipelines can match or 
surpass proprietary and ensemble-based systems on key pedagogical dimensions, 
with code and data available at \url{https://github.com/AIM-SCU/GRADE}.
\end{abstract}

\section{Introduction}

Large language models are increasingly deployed as AI tutors, yet evaluating 
whether they teach well requires more than factual correctness. Effective 
tutors must identify student mistakes, locate errors, provide meaningful 
guidance, and offer actionable next steps \cite{tack-piech-2022-ai, 
maurya-etal-2025-unifying}, but varied evaluation criteria across prior work 
make cross-system comparison difficult, motivating standardized automatic 
assessment \cite{tack-piech-2022-ai, tack-etal-2023-bea}.

\begin{figure}[h]
\includegraphics[width=\columnwidth]{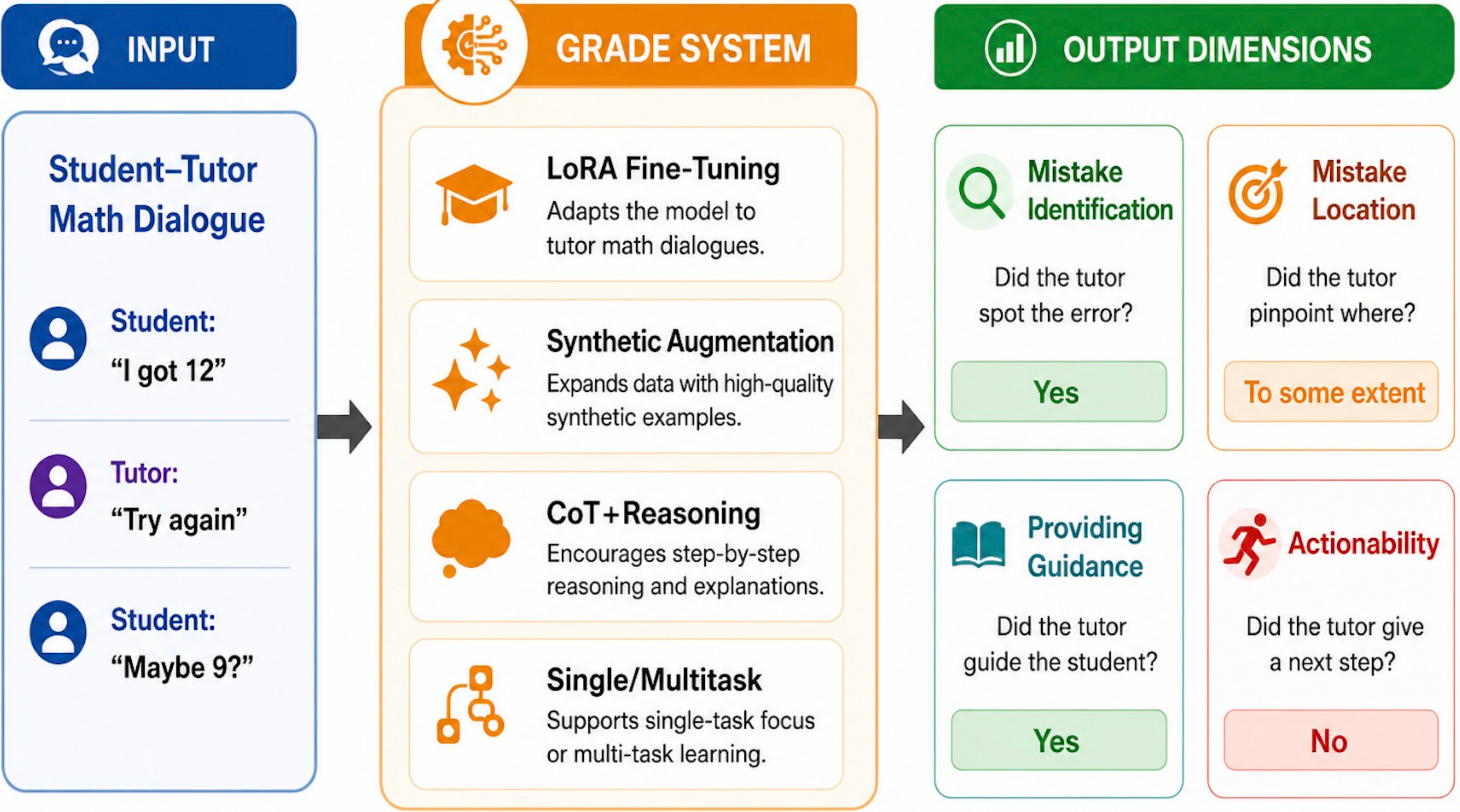}
\caption{Overview of the GRADE pipeline: a student-tutor math dialogue is 
processed through LoRA fine-tuning, synthetic augmentation, CoT+Reasoning, 
and single/multitask formulations, producing evaluations across four 
pedagogical dimensions.}
\label{fig:overview}
\end{figure}

The BEA 2025 Shared Task on Pedagogical Ability Assessment of AI-powered 
Tutors \cite{kochmar-etal-2025-findings} addresses this through a unified 
framework grounded in learning science \cite{maurya-etal-2025-unifying}, 
using math dialogues from MathDial \cite{macina-etal-2023-mathdial} and 
Bridge \cite{wang-etal-2024-bridging}. The task evaluates four dimensions: 
mistake identification, mistake location, providing guidance, and 
actionability. While parameter-efficient fine-tuning \cite{hu-etal-2022-lora} 
and synthetic augmentation improved performance, most work focused on 
individual dimensions and standard non-reasoning models, leaving reasoning, 
multitask learning, and efficiency tradeoffs underexplored 
\cite{dekmak-etal-2025-tutormind, kochmar-etal-2025-findings}.

In particular, the role of chain-of-thought reasoning \cite{wei-etal-2022-chain} 
remains underexplored across two stages: as a direct training and inference 
strategy, and as a tool for generating higher-quality synthetic data. Models 
such as Qwen3 \cite{qwen-team-2025-qwen3} support both thinking and 
non-thinking modes, enabling isolation of reasoning's contribution at each 
stage. Whether jointly modeling all four dimensions in a multitask formulation 
outperforms separate task-specific models also remains open. Finally, despite 
growing interest in sustainable NLP, the environmental cost of these choices 
has not been studied in this setting \cite{strubell-etal-2019-energy}.

We build on the strongest open-source BEA 2025 baseline and present GRADE 
(Figure~\ref{fig:overview}), conducting 120 experimental runs across model 
scale, training method, augmentation strategy, and task formulation. 
Understanding which modeling choices lead to reliable pedagogical evaluation 
is essential for building trustworthy AI tutoring systems and guiding future 
work in educational NLP. Our work is guided by three research questions:

\begin{itemize}
    \item[\textbf{RQ1}] How do model scale and fine-tuning strategy affect 
    pedagogical ability assessment in educational dialogue evaluation?
    
    \item[\textbf{RQ2}] Does chain-of-thought reasoning help more as a tool 
    for synthetic data generation, as a direct training and inference 
    strategy, or both?
    
    \item[\textbf{RQ3}] Does jointly training one model in a multitask 
    formulation across all four pedagogical dimensions yield more balanced 
    and competitive performance than independently trained single-task models?
\end{itemize}

Our contributions are threefold: (1) a systematic study across all four 
BEA 2025 dimensions covering model scale, training method, augmentation, 
and task formulation; (2) chain-of-thought reasoning analysis at both 
inference and data generation time, including self-verification filtering; 
and (3) the first carbon emissions analysis of pedagogical evaluation systems 
using CodeCarbon \cite{courty2024codecarbon}, with practical recommendations 
for compute-efficient model selection.

\section{Related Work}
\subsection{AI Tutor Evaluation in Educational Dialogues}

Evaluating AI tutor responses has become an important problem, but early work used varied criteria that made systems difficult to compare. \citet{tack-piech-2022-ai} proposed an early framework for measuring whether generative models could speak like teachers, understand students, and provide help, while the BEA 2023 Shared Task \cite{tack-etal-2023-bea} further emphasized response generation in educational dialogues. More recent work argues that evaluation should be grounded in learning science and move beyond surface level metrics \cite{jurenka-etal-2024-learnlm}.

\citet{maurya-etal-2025-unifying} addressed this gap by introducing a unified taxonomy for pedagogical ability assessment, which later formed the basis of the BEA 2025 Shared Task \cite{kochmar-etal-2025-findings}. The task operationalizes four dimensions, mistake identification, mistake location, providing guidance, and actionability, using educational math dialogues from MathDial \cite{macina-etal-2023-mathdial} and Bridge \cite{wang-etal-2024-bridging}. Our work builds on this shared task by extending evaluation across open-source models, reasoning-capable systems, and multitask training across all four dimensions.

\subsection{Fine-Tuning and Parameter-Efficient Adaptation of LLMs}

Fine-tuning is widely used for downstream NLP tasks, but updating all 
parameters is computationally expensive. LoRA \cite{hu-etal-2022-lora} 
reduces this cost by freezing pretrained weights and training small 
rank-decomposition modules, and became a common strategy in BEA 2025, 
where carefully tuned open-source models competed with larger proprietary 
systems \cite{kochmar-etal-2025-findings}.

Prior systems explored varied approaches: TutorMind 
\cite{dekmak-etal-2025-tutormind} combined LoRA with synthetic augmentation 
for minority classes; BJTU \cite{fan-etal-2025-bjtu} used task-aware prompt 
tuning; MSA \cite{hikal-etal-2025-msa} added disagreement-aware ensemble 
inference; bea-jh \cite{roh-bang-2025-beajh} applied Group Relative Policy 
Optimization with thinking-based rationales; BLCU-ICALL 
\cite{an-etal-2025-blcu} compared supervised fine-tuning, in-context 
learning, and reinforcement learning; and K-NLPers \cite{park-etal-2025-k} 
relied on GPT-4.1 prompting without fine-tuning. Together these confirm 
parameter-efficient fine-tuning as a strong baseline, though reasoning and 
task formulation remain underexplored across all four dimensions.

\subsection{Data Augmentation for Imbalanced Classification}

Class imbalance remains a major challenge in NLP classification, as majority labels can cause models to underperform on informative minority categories. Prior augmentation methods include rule based transformations, back translation, and model based generation \cite{feng-etal-2021-survey}. Recent work increasingly uses LLMs as data generators, and educational NLP studies show that open-source LLMs can provide useful feedback and training signal when guided appropriately \cite{koutcheme-etal-2024-opensource}.

In the BEA 2025 Shared Task, imbalance was especially severe, with the majority class covering nearly 78\% of annotations in mistake identification \cite{kochmar-etal-2025-findings}. Systems addressed this through synthetic data generation, oversampling, and class weighted losses. TutorMind \cite{dekmak-etal-2025-tutormind} generated minority class examples but noted label noise, while TBA \cite{gombert-etal-2025-tba} explored DARE-TIES model merging and NLIP \cite{saha-etal-2025-nlip} combined oversampling with multi-task learning. Our work builds on these efforts by using a reasoning-capable model for 
augmentation and adding a self-verification step to filter synthetic 
examples before training, targeting the most underrepresented minority 
labels across all four pedagogical dimensions.

\subsection{Reasoning and Multi-Task Learning in NLP}

Chain-of-thought prompting \cite{wei-etal-2022-chain} showed that intermediate reasoning can improve performance on complex tasks, motivating models with built-in reasoning capabilities. Qwen3 \cite{qwen-team-2025-qwen3} supports both thinking and non-thinking modes, enabling direct comparison between deliberate reasoning and standard inference. Alongside this, open-source models such as LLaMA 3 \cite{dubey-etal-2024-llama3}, Mistral \cite{jiang-etal-2023-mistral}, and Gemma 3 \cite{gemma-team-2025-gemma3} have made rigorous experimentation increasingly feasible without relying only on proprietary systems.

Multi-task learning offers a complementary direction by training one model across related objectives to exploit shared structure. In BEA 2025, most systems treated each pedagogical dimension separately, while TBA \cite{gombert-etal-2025-tba} showed that cross-dimension information can be useful through model merging. Our work extends this direction by directly comparing single-task and multi-task LoRA fine-tuning, while also testing whether chain-of-thought reasoning complements multi-task pedagogical evaluation.

\section{Dataset}

Following \cite{dekmak-etal-2025-tutormind}, we build on the BEA 2025 
TutorMind data for pedagogical evaluation of tutor responses. While prior 
work evaluated only a single dimension, a comprehensive assessment requires 
all four pedagogical dimensions jointly, motivating our extended setup. Each 
example is a tutor--student interaction with a task-specific label: 
\textit{Yes}, \textit{No}, or \textit{To some extent}. We evaluate four 
dimensions: \textit{Mistake Identification} (MI), \textit{Mistake Location} 
(ML), \textit{Providing Guidance} (PG), and \textit{Actionability} (ACT).

TutorMind focused only on MI, using 2,476 instances (1,980 train / 496 val) 
augmented with 200 minority-class examples. We extend this to all four 
dimensions and a multitask (MT) formulation predicting all labels jointly. 
For MI, splits are derived from the augmented release after removing 8 exact 
duplicates. For ML, PG, ACT, and MT, we retain task-valid examples, remove 
N/A targets and exact duplicates, producing the splits in 
Table~\ref{tab:dataset_summary}.

\begin{table}[t]
\centering
\small
\setlength{\tabcolsep}{4pt}
\begin{tabular}{lrrr}
\toprule
\textbf{Dataset} & \textbf{Train} & \textbf{Val} & \textbf{Total} \\
\midrule
TutorMind original MI & 1,980 & 496 & 2,476 \\
TutorMind augmented MI pool & 2,180 & 496 & 2,676 \\
Deduplicated augmented MI pool & -- & -- & 2,668 \\
\midrule
Our MI & 2,130 & 538 & 2,668 \\
Our ML & 1,974 & 494 & 2,468 \\
Our PG & 1,976 & 494 & 2,470 \\
Our Act & 1,974 & 494 & 2,468 \\
Our MT & 1,976 & 494 & 2,470 \\
\midrule
Our MI + Gen & 3,130 & 538 & 3,668 \\
Our ML + Gen & 2,974 & 494 & 3,468 \\
Our PG + Gen & 2,976 & 494 & 3,470 \\
Our Act + Gen & 2,974 & 494 & 3,468 \\
Our MT + Gen & 2,976 & 494 & 3,470 \\
\midrule
Our MI + Gen+Verify & 3,130 & 538 & 3,668 \\
Our ML + Gen+Verify & 2,974 & 494 & 3,468 \\
Our PG + Gen+Verify & 2,976 & 494 & 3,470 \\
Our Act + Gen+Verify & 2,974 & 494 & 3,468 \\
Our MT + Gen+Verify & 2,607 & 494 & 3,101 \\
\bottomrule
\end{tabular}
\caption{Summary of Proposed Dataset Splits}
\label{tab:dataset_summary}
\vspace{-2.0em}
\end{table}

To address class imbalance, we construct two augmented variants with 
Qwen3-14B: \textit{Qwen3 Gen}, where synthetic minority-class examples are 
used directly, and \textit{Qwen3 Gen+Verify}, where generated examples are 
first verified by the same model. Each Gen dataset adds 1,000 synthetic 
examples (500/500 split per task), while MT + Gen+Verify contains 631 
examples for reasons detailed in Section~\ref{sec:augmentation}.

\section{Methodology}
\subsection{Task Formulation}

Each pedagogical dimension is treated as an independent three-class 
classification problem over student-tutor math dialogues. Given the full 
conversational context and a tutor response, a model assigns one of three 
labels: \textit{Yes}, \textit{No}, or \textit{To some extent}. This 
three-way distinction is more demanding than binary classification, as the 
intermediate class captures responses that partially satisfy a pedagogical 
criterion without fully meeting it, and accounts for the bulk of 
minority-class examples.

For single-task experiments, each model is trained and evaluated on one 
dimension at a time, producing a single \texttt{Evaluation:} line. For the 
multitask setting, all four dimensions are evaluated jointly within a single 
forward pass, requiring four structured output lines simultaneously. This 
joint formulation tests whether a single model can internalize 
cross-dimension relationships without task-specific specialization. Prompts 
for both settings are in Tables~\ref{tab:prompts-standard} 
and~\ref{tab:prompts-cot}.

\subsection{Models and Experimental Design}

We evaluate five open-source instruction-tuned models: LLaMA-3.1-8B 
\cite{dubey-etal-2024-llama3}, Mistral-7B \cite{jiang-etal-2023-mistral}, 
Qwen3-14B \cite{qwen-team-2025-qwen3}, Gemma3-12B 
\cite{gemma-team-2025-gemma3}, and Gemma3-27B 
\cite{gemma-team-2025-gemma3}. LLaMA-3.1-8B and Mistral-7B replicate the 
TutorMind fine-tuning setup \cite{dekmak-etal-2025-tutormind}, while larger 
models test the effect of scale. Qwen3-14B additionally studies 
chain-of-thought reasoning via its native thinking mode.

Our design, summarized in Table~\ref{tab:experiments}, spans 120 runs 
across model scale, training method, augmentation strategy, reasoning mode, 
and task formulation. We compare zero-shot inference and LoRA fine-tuning 
\cite{hu-etal-2022-lora}, isolate chain-of-thought reasoning by toggling 
Qwen3-14B between Think OFF and Think ON, evaluate generated versus 
self-verified augmentation, and compare single-task with multitask models.

\subsection{Fine-Tuning and Reasoning}

All fine-tuning uses LoRA \cite{hu-etal-2022-lora} with Unsloth 
\cite{han-etal-2023-unsloth}, applying adapters to all attention and 
feed-forward projection matrices with rank $r=16$, scaling factor 
$\alpha=16$, and no dropout. Models are trained in bfloat16, except 
Gemma3-27B \cite{gemma-team-2025-gemma3}, which uses 8-bit quantization 
due to compute constraints. Training runs for 3 epochs with batch size 2, 
gradient accumulation over 8 steps, AdamW \cite{loshchilov-hutter-2019-adamw}, 
learning rate $2 \times 10^{-4}$, 5 warmup steps, and weight decay 0.01. 
Inference uses greedy decoding with at most 64 new tokens.

\begin{table}[h]
\centering
\footnotesize
\setlength{\tabcolsep}{4pt}
\renewcommand{\arraystretch}{0.85}
\begin{tabular}{l p{4.2cm} r}
\toprule
\textbf{Experiment} & \textbf{Configuration} & \textbf{Runs} \\
\midrule
\multirow{2}{*}{Baseline} & 2 models, Zero-shot, No Aug & 10 \\
                          & 2 models, LoRA, No Aug & 10 \\
\midrule
\multirow{2}{*}{\shortstack[l]{New Models \\ (Ours)}} & 
    3 models, Zero-shot, No Aug & 15 \\
  & 3 models, LoRA, No Aug & 15 \\
\midrule
\multirow{2}{*}{\shortstack[l]{Reasoning \\ (Ours)}} & 
    Qwen3, Zero-shot, Think ON & 5 \\
  & Qwen3, LoRA, Think ON & 5 \\
\midrule
\multirow{2}{*}{\shortstack[l]{Augmentation \\ (Ours)}} & 
    5 models, LoRA, Generate & 25 \\
  & 5 models, LoRA, Gen+Verify & 25 \\
\midrule
\multirow{2}{*}{\shortstack[l]{Combined \\ (Ours)}} & 
    Qwen3, LoRA, Think ON, Gen & 5 \\
  & Qwen3, LoRA, Think ON, Gen+Verify & 5 \\
\midrule
\textbf{Total} & & \textbf{120} \\
\bottomrule
\end{tabular}
\caption{Summary of experimental configurations. Each run covers all five 
classification tasks.}
\label{tab:experiments}
\vspace{-1.40em}
\end{table}

Chain-of-thought reasoning is studied via Qwen3-14B 
\cite{qwen-team-2025-qwen3}, which supports native thinking mode. Think ON 
is enabled through \texttt{enable\_thinking=True} with a reasoning-augmented 
prompt; Think OFF uses the standard prompt. This controlled toggle isolates 
reasoning's effect on classification. Full prompts are in 
Tables~\ref{tab:prompts-standard} and~\ref{tab:prompts-cot}.

\subsection{Reasoning-Guided Data Augmentation}
\label{sec:augmentation}

The BEA 2025 dataset is highly imbalanced, so we use Qwen3-14B with 
CoT+Reasoning to generate synthetic examples for minority labels 
(\textit{No} and \textit{To some extent}). For each task, the model 
receives a real student-tutor conversation and generates a one-sentence 
tutor response matching the target label. We compare \textit{Gen}, which 
uses generated examples directly, with \textit{Gen+Verify} (Gen+Ver.\ in 
tables), where Qwen3-14B retains only examples whose self-predicted label 
matches the intended minority class label. Full prompts are in 
Tables~\ref{tab:prompts-augmentation} and~\ref{tab:prompts-verification}.

Each Gen dataset adds 1,000 synthetic examples split evenly between 
\textit{No} and \textit{To some extent}. Gen+Verify uses the same split, 
while MT + Gen+Verify contains only 631 examples: the verification step 
requires synthetic responses to satisfy all four dimensions simultaneously 
at the \textit{To some extent} level, a constraint the model rarely 
fulfills, resulting in high rejection rates. This corroborates that 
self-verification introduces substantial overhead without consistent gains.

\section{Evaluation \& Results}

\subsection{Evaluation Metrics}
We evaluate all systems using strict and lenient macro-averaged F1 and 
accuracy, following the official BEA 2025 protocol 
\cite{kochmar-etal-2025-findings}. Strict evaluation treats \textit{Yes}, 
\textit{No}, and \textit{To some extent} as three separate classes, while 
lenient evaluation merges \textit{Yes} and \textit{To some extent}.
We use strict macro-averaged F1 as the primary metric as accuracy can be 
inflated under strong class imbalance \cite{dekmak-etal-2025-tutormind}. 
Statistical significance is assessed using 95\% confidence intervals via 
bootstrap resampling. Lenient F1 results are reported in 
Appendix~\ref{sec:lenient-results}.

\subsection{Baseline Performance Without Augmentation}

\paragraph{Zero Shot Analysis.}
Figure~\ref{fig:zeroshot-single} reports strict F1 scores for the 
single-task zero-shot setting. MI is the strongest dimension across all 
models, with Qwen3-14B achieving the highest score of 0.622. ML is 
consistently the weakest, with Mistral-7B scoring as low as 0.288, 
confirming that localizing an error is substantially harder than detecting 
its presence. Gemma3-12B and Gemma3-27B (8-bit) are the most balanced 
overall, maintaining competitive scores across all four dimensions without 
any fine-tuning signal.

\begin{center}
\includegraphics[width=0.95\columnwidth]{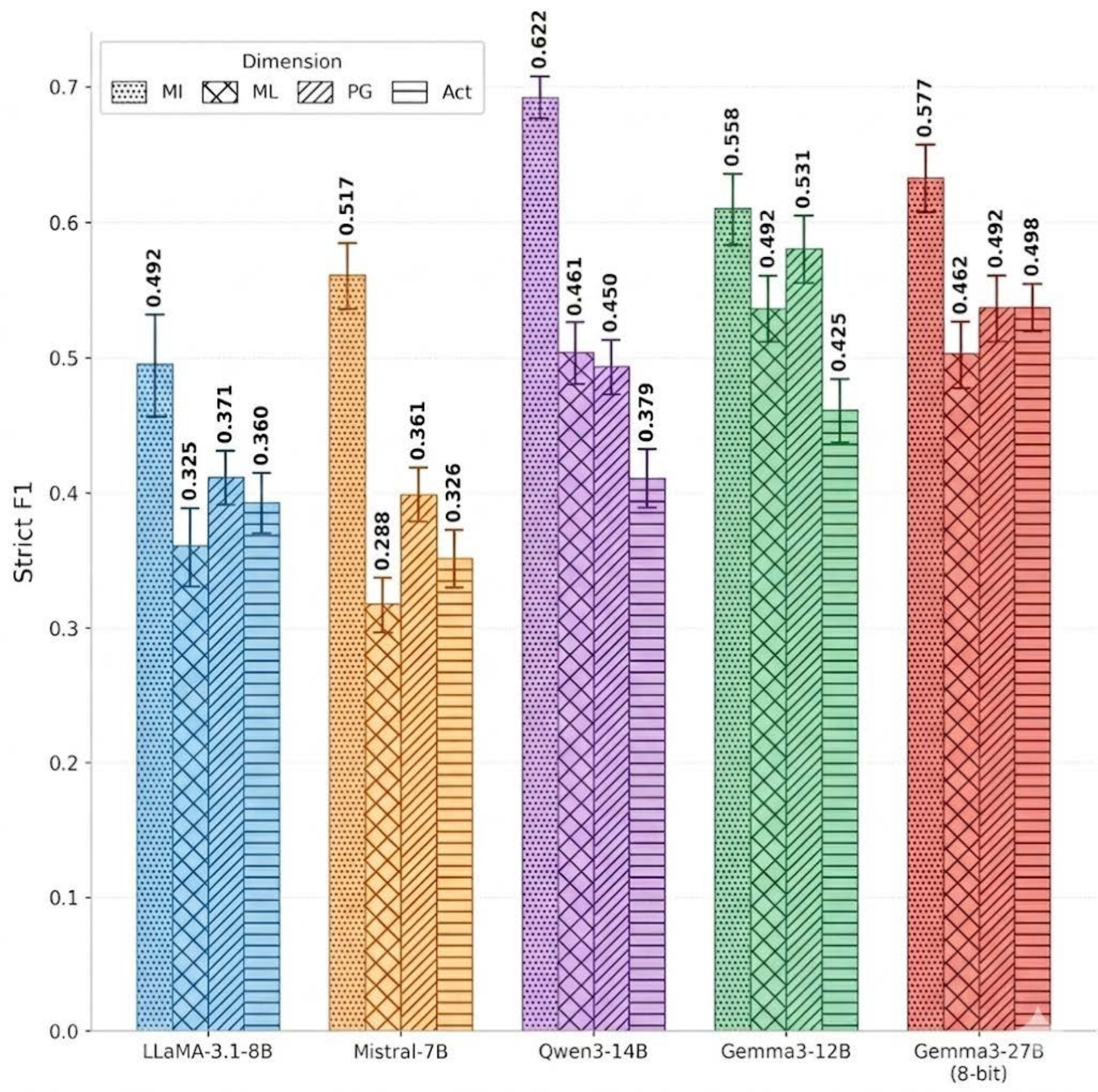}
\captionof{figure}{Single task zero shot strict F1 scores without 
augmentation across all five models and four pedagogical dimensions.}
\label{fig:zeroshot-single}
\vspace{-0.5em}
\end{center}

Figure~\ref{fig:zeroshot-mt} reports multitask zero-shot results. Joint 
prediction lowers performance for most models, with the largest drop for 
Mistral-7B, whose MI score falls from 0.517 to 0.329. Gemma3-27B (8-bit) 
is the most robust, retaining the highest MI score at 0.515 and remaining 
balanced across PG and Act, suggesting larger models handle joint 
prediction better under zero-shot inference.

\paragraph{LoRA Analysis.}
Figure~\ref{fig:lora-single} reports strict F1 after LoRA fine-tuning in 
the single-task setting. Gains are not uniform: Gemma3-12B achieves the 
strongest result at 0.750 on MI with competitive scores on PG and Act. 
Qwen3-14B improves substantially to 0.717 on MI, and LLaMA-3.1-8B shows 
clear gains across dimensions. Mistral-7B remains weak, and Gemma3-27B 
shows unstable behavior on MI and ML, suggesting 8-bit quantization limits 
its benefit from parameter-efficient fine-tuning.

\begin{center}
\includegraphics[width=0.95\columnwidth]{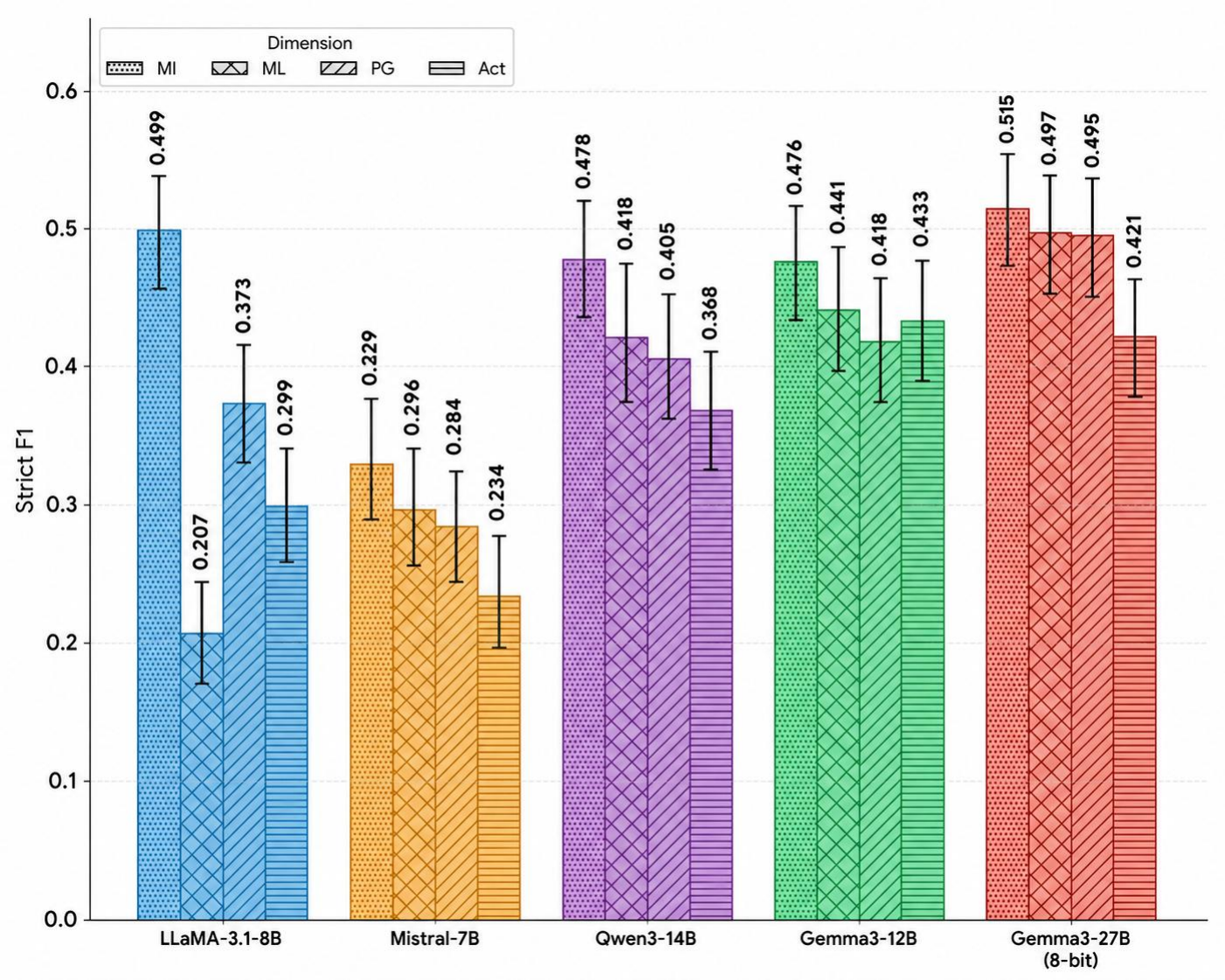}
\captionof{figure}{Multitask zero shot strict F1 scores without 
augmentation across all five models and four pedagogical dimensions.}
\label{fig:zeroshot-mt}
\end{center}

\begin{center}
\includegraphics[width=0.95\columnwidth]{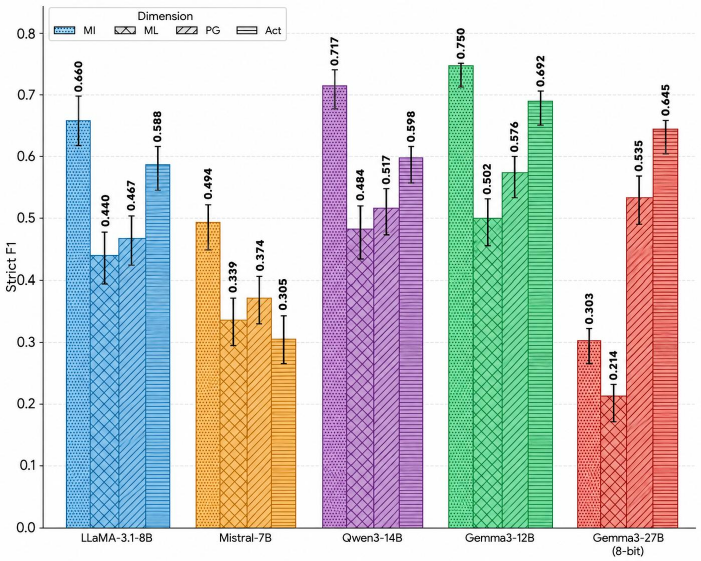}
\captionof{figure}{Single task LoRA strict F1 scores without augmentation across all five models and four pedagogical dimensions.}
\label{fig:lora-single}
\end{center}

Figure~\ref{fig:lora-mt} shows the multitask setting reverses this 
pattern. Gemma3-27B becomes the strongest multitask model, reaching 0.598 
on MI and 0.644 on Act with balanced performance across PG and ML, while 
Gemma3-12B loses its single-task advantage. Overall, LoRA is an effective 
baseline, but its benefit depends on model scale, quantization, and task 
formulation: single-task training favors Gemma3-12B, while multitask 
training better suits Gemma3-27B, with larger models benefiting most from 
the joint formulation.

\subsection{Effect of Data Augmentation \& Verification}

Table~\ref{tab:augmentation-results} shows that augmentation has highly 
model-dependent effects. The clearest gain appears for Gemma3-27B (8-bit) 
in the single-task setting, where MI rises from 0.30 to 0.77 with Qwen3 
Gen+Verify, suggesting augmented data substantially recovers performance 
under the constrained 8-bit LoRA setup. Mistral-7B also benefits, 
especially in multitask evaluation, where MI improves from 0.31 to 0.48. 
In contrast, higher-performing baseline models such as Gemma3-12B, Qwen3-14B, and 
LLaMA-3.1-8B show limited gains or small regressions, suggesting their 
learning capacity on the original data is already saturated. Synthetic 
minority-class examples are thus most beneficial when a model has not yet 
reached its learning capacity, rather than as a universal strategy. 
Differences between Gen and Gen+Verify are small and inconsistent, 
indicating no reliable advantage from verification, consistent with the 
high rejection rates in MT Gen+Verify detailed in 
Section~\ref{sec:augmentation}.

\begin{center}
\includegraphics[width=0.95\columnwidth]{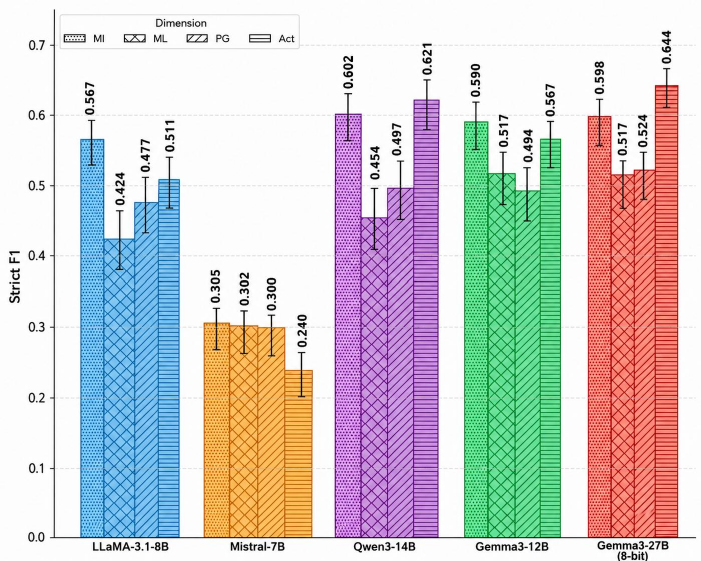}
\captionof{figure}{Multitask LoRA strict F1 scores without augmentation across all five models and four pedagogical dimensions.}
\label{fig:lora-mt}
\end{center}

\begin{table*}[h]
\centering
\scriptsize
\setlength{\tabcolsep}{2.7pt}
\renewcommand{\arraystretch}{1.08}
\resizebox{\textwidth}{!}{
\begin{tabular}{llcccccccccccc}
\toprule
\textbf{Setting} & \textbf{Model}
& \multicolumn{4}{c}{\textbf{No Aug}}
& \multicolumn{4}{c}{\textbf{Qwen3 Gen}}
& \multicolumn{4}{c}{\textbf{Qwen3 Gen+Verify}} \\
\cmidrule(lr){3-6} \cmidrule(lr){7-10} \cmidrule(lr){11-14}
& & \textbf{MI} & \textbf{ML} & \textbf{PG} & \textbf{Act}
& \textbf{MI} & \textbf{ML} & \textbf{PG} & \textbf{Act}
& \textbf{MI} & \textbf{ML} & \textbf{PG} & \textbf{Act} \\
\midrule
\multirow{5}{*}{Single}
& LLaMA-3.1-8B & 0.66 & 0.44 & 0.47 & 0.59 & 0.63 & 0.44 
    & 0.48 & 0.55 & 0.63 & 0.45 & 0.48 & 0.55 \\
& Mistral-7B   & 0.49 & 0.34 & 0.37 & 0.31 & 0.51 & 0.34 
    & 0.40 & 0.43* & 0.50 & 0.37 & 0.40 & 0.47* \\
& Qwen3-14B    & 0.72 & 0.48 & 0.52 & 0.60 & 0.70 & 0.47 
    & 0.50 & 0.57 & 0.70 & 0.48 & 0.53 & 0.61 \\
& Gemma3-12B   & \textbf{0.75} & \textbf{0.50} & \textbf{0.58} 
    & \textbf{0.69} & 0.73 & \textbf{0.53} & 0.55 & 0.58\dag 
    & 0.73 & \textbf{0.51} & \textbf{0.58} & \textbf{0.64} \\
& Gemma3-27B (8-bit) & 0.30 & 0.21 & 0.54 & 0.65 
    & \textbf{0.75*} & 0.51* & \textbf{0.57} & \textbf{0.58}\dag 
    & \textbf{0.77*} & \textbf{0.51*} & 0.56 & 0.60 \\
\midrule
\multirow{5}{*}{MT}
& LLaMA-3.1-8B & 0.57 & 0.42 & 0.48 & 0.51 & 0.56 & 0.42 
    & 0.49 & 0.48 & 0.56 & 0.43 & 0.48 & 0.48 \\
& Mistral-7B   & 0.31 & 0.30 & 0.30 & 0.24 & 0.48* & 0.45* 
    & 0.50* & 0.48* & 0.48* & 0.48* & 0.50* & 0.49* \\
& Qwen3-14B    & \textbf{0.60} & 0.45 & 0.50 & 0.62 & 0.55 
    & 0.49 & \textbf{0.53} & 0.57 & \textbf{0.62} & 0.47 
    & 0.51 & 0.58 \\
& Gemma3-12B   & 0.59 & 0.52 & 0.49 & 0.57 & 0.57 
    & \textbf{0.51} & 0.49 & \textbf{0.60} & 0.55 & 0.50 
    & 0.49 & 0.57 \\
& Gemma3-27B (8-bit) & 0.60 & \textbf{0.52} & \textbf{0.52} 
    & \textbf{0.64} & \textbf{0.60} & 0.50 & 0.53 & 0.58 
    & 0.62 & \textbf{0.51} & \textbf{0.52} & \textbf{0.59} \\
\bottomrule
\end{tabular}
}
\caption{Strict F1 scores for LoRA models with Think OFF under no 
augmentation, Qwen3 generated augmentation, and Qwen3 generated plus 
verified augmentation. Bold: strongest per setting and augmentation group. 
* significant improvement over No Aug baseline; \dag\ significant decrease .}
\label{tab:augmentation-results}
\end{table*}

\subsection{Chain-of-Thought and Reasoning Analysis}

Tables~\ref{tab:qwen-thinking-noaug} and~\ref{tab:qwen-thinking-aug} show 
that CoT+Reasoning has a stage dependent effect. Without augmentation, Think ON weakens performance: the effect is negligible under zero shot 
inference, but severe under LoRA fine tuning, where MI falls from 0.72 to 
0.22 with similar collapse across dimensions. This is not due to token 
budget, since each instance receives 1,024 thinking tokens. Instead, 
inspection shows an empty reasoning chain, while the model generates free 
form mathematical solutions rather than structured labels. This suggests 
that failure comes from the interaction between LoRA fine tuning and 
thinking mode, rather than thinking mode alone, since zero shot Think ON 
has negligible unknown rates of $\sim$1--2\%. The result highlights a 
broader concern for reasoning capable models under parameter efficient 
fine tuning.

\begin{table}[h]
\centering
\footnotesize
\setlength{\tabcolsep}{2.5pt}
\renewcommand{\arraystretch}{1.08}
\begin{tabular}{lcccc}
\toprule
\textbf{Setting} & \textbf{MI} & \textbf{ML} & \textbf{PG} & \textbf{Act} \\
\midrule
ZS, Think OFF & \textbf{0.62} & \textbf{0.46} & \textbf{0.45} & \textbf{0.38} \\
ZS, Think ON  & 0.52\dag & 0.46 & 0.45 & 0.36 \\
LoRA, Think OFF & \textbf{0.72} & \textbf{0.48} & \textbf{0.52} & \textbf{0.60} \\
LoRA, Think ON  & 0.22\dag & 0.28\dag & 0.24\dag & 0.29\dag \\
\bottomrule
\end{tabular}
\caption{Strict F1 for Qwen3-14B with Think OFF and Think ON without 
augmentation. \dag\ significant decrease relative to Think OFF.}
\label{tab:qwen-thinking-noaug}
\end{table}

With augmented data, performance recovers sharply 
(Table~\ref{tab:qwen-thinking-aug}): single-task MI reaches 0.72 and 
multitask MI reaches 0.62, with zero unparseable outputs under both 
strategies. Gen and Gen+Verify remain close with no consistent winner. 
Overall, CoT+Reasoning is more effective as a data enrichment mechanism 
than a direct inference strategy --- standard inference should be preferred 
for classification, while reasoning mode is best reserved for data 
construction.

\begin{table}[h]
\centering
\footnotesize
\setlength{\tabcolsep}{2.5pt}
\renewcommand{\arraystretch}{1.08}
\begin{tabular}{llcccc}
\toprule
\textbf{Set.} & \textbf{Aug.} & \textbf{MI} & \textbf{ML} 
    & \textbf{PG} & \textbf{Act} \\
\midrule
\multirow{3}{*}{ST}
& No Aug    & 0.22 & 0.28 & 0.24 & 0.29 \\
& Gen       & \textbf{0.72}* & 0.50* & \textbf{0.51}* & 0.62* \\
& Gen+Ver.  & 0.71* & \textbf{0.50}* & 0.49* & \textbf{0.64}* \\
\midrule
\multirow{2}{*}{MT}
& Gen       & \textbf{0.62}* & \textbf{0.50}* & \textbf{0.52}* & 0.56* \\
& Gen+Ver.  & 0.57* & 0.48* & 0.51* & \textbf{0.59}* \\
\bottomrule
\end{tabular}
\caption{Strict F1 for Qwen3-14B with LoRA and Think ON across 
augmentation strategies. No Aug MT omitted (outputs collapsed to Unknown). 
* significant improvement over No Aug.}
\label{tab:qwen-thinking-aug}
\end{table}

\subsection{Comparison with Published Benchmarks}

Table~\ref{tab:published-comparison} compares our strongest configurations 
with prior BEA 2025 systems on the same development set and strict macro F1 
protocol \cite{an-etal-2025-blcu, dekmak-etal-2025-tutormind, 
roh-bang-2025-beajh, hikal-etal-2025-msa}. Gemini 2.5 Pro is strongest on 
ML (0.68) and PG (0.67), while our Gemma3 27B with GenVer reaches the best 
MI score (0.77). Gemma3 12B NoAug matches MSA on Act (0.69). Overall, open 
source LoRA can exceed closed source and ensemble based systems on MI while 
remaining broadly competitive across the other pedagogical dimensions.

\begin{table}[h]
\centering
\footnotesize
\setlength{\tabcolsep}{3.2pt}
\renewcommand{\arraystretch}{1.08}
\begin{tabular}{llcccc}
\toprule
\textbf{Paper} & \textbf{Method} & \textbf{MI} & \textbf{ML} 
    & \textbf{PG} & \textbf{Act} \\
\midrule
\href{https://aclanthology.org/2025.bea-1.84/}{BLCU} 
    & GPT4o ICL & 0.46 & 0.45 & 0.49 & 0.44 \\
\href{https://aclanthology.org/2025.bea-1.84/}{BLCU} 
    & o3-mini ICL & 0.43 & 0.47 & 0.49 & 0.45 \\
\href{https://aclanthology.org/2025.bea-1.84/}{BLCU} 
    & Gemini 2.5 Pro ICL & 0.62 & \textbf{0.68} & \textbf{0.67} & 0.56 \\
\href{https://aclanthology.org/2025.bea-1.84/}{BLCU} 
    & Grok ICL & 0.48 & 0.39 & 0.43 & 0.42 \\
\href{https://aclanthology.org/2025.bea-1.84/}{BLCU} 
    & DeepSeek ICL & 0.51 & 0.51 & 0.54 & 0.47 \\
\href{https://aclanthology.org/2025.bea-1.84/}{BLCU} 
    & Claude ICL & 0.45 & 0.49 & 0.63 & 0.43 \\
\midrule
\href{https://aclanthology.org/2025.bea-1.96/}{TutorMind} 
    & GPT4o Aug & 0.70 & -- & -- & -- \\
\href{https://aclanthology.org/2025.bea-1.80/}{bea-jh} 
    & GLM SFT & 0.72 & 0.48 & 0.59 & 0.54 \\
\href{https://aclanthology.org/2025.bea-1.80/}{bea-jh} 
    & GLM GRPO & 0.56 & 0.57 & 0.57 & 0.66 \\
\href{https://aclanthology.org/2025.bea-1.95/}{MSA} 
    & Mathstral Ens & 0.72 & 0.57 & 0.58 & \textbf{0.69} \\
\midrule
Ours & G3-12B NoAug & \textbf{0.75}* & 0.50 & \textbf{0.58} 
    & \textbf{0.69} \\
Ours & G3-12B Gen   & \textbf{0.74}* & \textbf{0.53} & 0.55 & 0.58 \\
Ours & G3-27B GenVer & \textbf{0.77}* & 0.51 & 0.57 & \textbf{0.60} \\
\bottomrule
\end{tabular}
\caption{Strict F1 comparison against prior BEA 2025 systems. * 
significant improvement over the best prior systems.}
\label{tab:published-comparison}
\end{table}

\subsection{Carbon Emission Analysis}
As NLP systems scale, environmental cost becomes essential for sustainable 
research \cite{strubell-etal-2019-energy}. We track emissions using 
CodeCarbon v3.2.6 \cite{courty2024codecarbon} on an NVIDIA L40S (GPU via 
NVML, CPU via TDP, RAM). Figures~\ref{fig:carbon-augmentation} 
and~\ref{fig:carbon-thinking} show carbon cost is shaped by model choice, 
augmentation, and reasoning mode. Emissions vary more strongly by base 
model than augmentation alone: Mistral 7B and LLaMA 3.1 8B remain 
carbon efficient, while Gemma3 27B becomes the dominant contributor once 
augmentation is introduced. Qwen3 14B shows a sharper increase under 
augmentation; Gemma3 12B increases comparatively little, making it a 
strong middle cost option. Think ON substantially increases emissions, 
especially in the multitask LoRA setting, and is only justified for data 
construction given no consistent classification benefit. Pure data 
generation costs approximately 2.0 kg CO$_2$ for Gen and 4.2 kg CO$_2$ 
for Gen+Verify, confirming verification roughly doubles the augmentation 
footprint with limited performance gains.

\section{Lessons Learned}
\begin{enumerate}
    \item \textbf{Model choice depends on task formulation.} Gemma3-12B is 
    strongest for single-task settings, while Gemma3-27B (8-bit) is most 
    consistent in multitask settings, where larger capacity outweighs 
    quantization costs (Figures~\ref{fig:lora-single} 
    and~\ref{fig:lora-mt}). Single-task and multitask objectives favor 
    different model capacities.

    \item \textbf{Pedagogical dimensions differ in difficulty.} MI is the 
    easiest and most learnable; ML remains hardest despite fine-tuning, 
    augmentation, and scale, as pinpointing error location in multi-step 
    solutions is fundamentally harder than detecting or describing the error. Act starts weak under zero-shot but improves substantially with LoRA, 
    while PG stays stable in the middle 
    (Figures~\ref{fig:zeroshot-single}--\ref{fig:lora-mt}).

    \item \textbf{Multitask training improves balance.} Single-task training 
    peaks on individual dimensions, but multitask training yields more 
    balanced predictions by sharing signal across labels 
    (Figures~\ref{fig:lora-single} and~\ref{fig:lora-mt}), promoting 
    label consistency at the cost of peak performance on individual 
    objectives.

    \item \textbf{Augmentation helps selectively.} It strongly benefits 
    constrained models such as Gemma3-27B (8-bit) and Mistral-7B, but 
    yields limited gains for stronger models like Gemma3-12B and Qwen3-14B 
    that already saturate the training signal 
    (Table~\ref{tab:augmentation-results}). Augmentation is most valuable 
    when model capacity is constrained relative to the data distribution.

    \item \textbf{Verification doubles cost without consistent gains.} Gen 
    and GenVer perform similarly, yet GenVer raises generation carbon cost 
    from 2.0 to 4.2 kg CO$_2$ 
    (Table~\ref{tab:augmentation-results}), making self-verification an 
    unnecessary overhead that does not generalize as a reliable quality 
    control mechanism.

    \item \textbf{CoT+Reasoning is better for data generation than 
    classification.} Think ON causes collapse under LoRA without 
    augmentation, but recovers performance when used to generate data 
    (Tables~\ref{tab:qwen-thinking-noaug} and~\ref{tab:qwen-thinking-aug}). 
    Reasoning mode is better exploited as a data construction tool than as 
    a direct classification strategy under parameter-efficient fine-tuning.

    \item \textbf{Carbon cost shapes practical recommendations.} Larger 
    models and Think ON introduce substantial emissions 
    (Figures~\ref{fig:carbon-augmentation} 
    and~\ref{fig:carbon-thinking}). Gemma3 12B has the smallest 
    emissions spike across augmentation conditions, making it the 
    strongest single task choice when both performance and efficiency 
    matter. These findings highlight the need for carbon aware model 
    selection in sustainable NLP research.

\end{enumerate}

\begin{center}
\includegraphics[width=0.95\columnwidth]{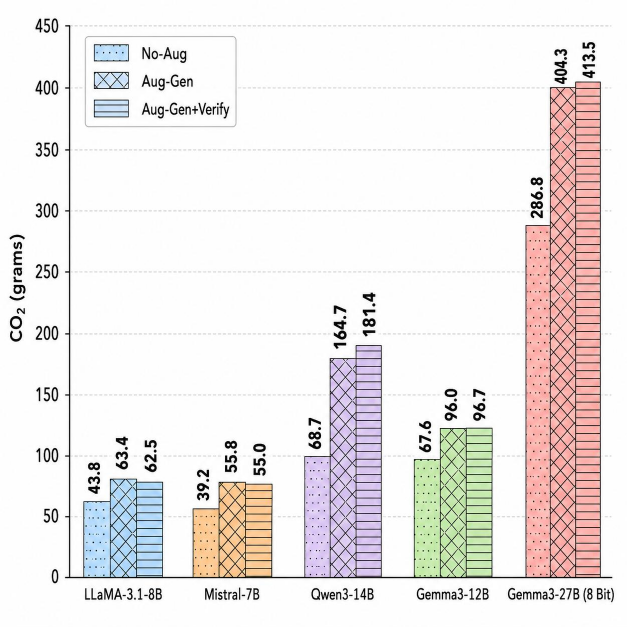}
\captionof{figure}{Carbon emissions for LoRA fine-tuning across models 
and augmentation strategies.}
\label{fig:carbon-augmentation}
\end{center}
\begin{center}
\includegraphics[width=0.95\columnwidth]{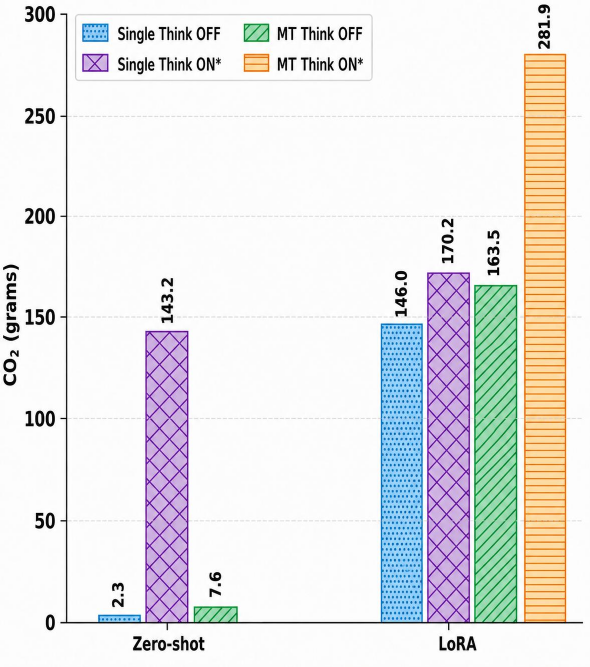}
\captionof{figure}{Carbon emissions for Qwen3-14B under different task 
and reasoning settings.}
\label{fig:carbon-thinking}
\end{center}

\section{Conclusion}

We present a systematic study of open source models for pedagogical ability 
assessment in AI tutor responses. Across 120 experimental runs covering 
zero shot inference, LoRA fine tuning, synthetic augmentation, 
CoT+Reasoning, and single task versus multitask formulations, we show that 
Gemma3 12B is strongest for single task evaluation, Gemma3 27B 
(8 bit) is more reliable in multitask settings, and open source LoRA 
pipelines can match or surpass proprietary and ensemble based systems on 
key pedagogical dimensions.

These findings offer practical guidance for building automatic evaluation 
systems: augmentation, reasoning, and model scale should be used 
selectively. Augmentation helps constrained models more than those already 
saturating the training signal, Gen+Ver does not consistently outperform Gen 
despite higher cost, and CoT+Reasoning is more useful for data generation 
than direct classification. LoRA fine tuning can also interfere with instruction following under thinking 
mode, raising broader concerns for reasoning capable models. Carbon analysis shows that stronger 
performance can carry substantial environmental cost. Future work should 
explore multilingual data, task weighting, label dependency modeling, and 
alternative reasoning capable models for augmentation beyond Qwen3. Code 
and datasets are available at \url{https://github.com/AIM-SCU/GRADE}.

\section{Limitations}

\paragraph{Dataset size.}
Although this work expands the original \cite{dekmak-etal-2025-tutormind} setting across all four pedagogical dimensions, the dataset remains relatively small for training and evaluating robust LLM based judges. More work is needed from the community to grow this benchmark so that both LLM judges and fine tuned models can learn from broader and more diverse examples.

\paragraph{Using only Qwen3 for augmentation and reasoning.}
Our augmentation and CoT+Reasoning experiments are centered on Qwen3 14B 
because it supports both synthetic generation and the Think ON and Think 
OFF setup used in this study. Since evaluation is performed on untouched 
validation data, the improvements still reflect genuine downstream gains 
rather than direct contamination. However, using a single model for both 
generation and reasoning limits how broadly these findings can be 
generalized. Future work should test whether similar gains hold with other 
reasoning capable models, such as DeepSeek R1\cite{deepseek-ai-2025-r1}, and with alternative 
augmentation sources or independent verifier models.

\paragraph{Quantized Gemma3-27B setting.}
Gemma3-27B is evaluated with 8-bit quantization due to compute constraints. This makes the 27B experiments practical, but it is not a clean full precision comparison against smaller models trained without quantization.

\paragraph{Mistake Location remains challenging.}
Mistake Location remains the most difficult dimension across our experiments. Its consistently lower scores suggest that the current prompts, training setup, and augmentation strategy still do not fully capture fine grained error localization in student solutions.

\paragraph{Single domain generalizability.}
Our findings are grounded in math tutoring dialogues 
(MathDial and Bridge). Whether these conclusions 
transfer to other educational domains, such as 
language learning or science tutoring, where error 
types and pedagogical strategies differ substantially, 
remains an open question for future work.

\bibliography{custom}

@inproceedings{tack-piech-2022-ai,
  title     = {The {AI} Teacher Test: Measuring the Pedagogical Ability of Blender and {GPT}-3 in Educational Dialogues},
  author    = {Tack, Ana{\"i}s and Piech, Chris},
  booktitle = {Proceedings of the 15th International Conference on Educational Data Mining},
  pages     = {522--529},
  year      = {2022},
  url       = {https://web.stanford.edu/~cpiech/bio/papers/aiteachertest.pdf}
}

@inproceedings{tack-etal-2023-bea,
  title     = {The {BEA} 2023 Shared Task on Generating {AI} Teacher Responses in Educational Dialogues},
  author    = {Tack, Ana{\"i}s and Kochmar, Ekaterina and Yuan, Zheng and Bibauw, Serge and Piech, Chris},
  booktitle = {Proceedings of the 18th Workshop on Innovative Use of {NLP} for Building Educational Applications},
  pages     = {785--795},
  year      = {2023},
  url       = {https://aclanthology.org/2023.bea-1.64}
}

@inproceedings{maurya-etal-2025-unifying,
  title     = {Unifying {AI} Tutor Evaluation: An Evaluation Taxonomy for Pedagogical Ability Assessment of {LLM}-Powered {AI} Tutors},
  author    = {Maurya, Kaushal Kumar and Srivatsa, KV Aditya and Petukhova, Kseniia and Kochmar, Ekaterina},
  booktitle = {Proceedings of the 2025 Conference of the Nations of the Americas Chapter of the Association for Computational Linguistics: Human Language Technologies},
  pages     = {1234--1251},
  year      = {2025},
  url       = {https://aclanthology.org/2025.naacl-long.57}
}

@inproceedings{kochmar-etal-2025-findings,
  title     = {Findings of the {BEA} 2025 Shared Task on Pedagogical Ability Assessment of {AI}-powered Tutors},
  author    = {Kochmar, Ekaterina and Maurya, Kaushal Kumar and Petukhova, Kseniia and Srivatsa, KV Aditya and Tack, Ana{\"i}s and Vasselli, Justin},
  booktitle = {Proceedings of the 20th Workshop on Innovative Use of {NLP} for Building Educational Applications},
  pages     = {1011--1033},
  year      = {2025},
  url       = {https://aclanthology.org/2025.bea-1.77}
}

@inproceedings{macina-etal-2023-mathdial,
  title     = {{MathDial}: A Dialogue Tutoring Dataset with Rich Pedagogical Properties Grounded in Math Reasoning Problems},
  author    = {Macina, Jakub and Daheim, Nico and Chowdhury, Sankalan and Sinha, Tanmay and Kapur, Manu and Gurevych, Iryna and Sachan, Mrinmaya},
  booktitle = {Findings of the Association for Computational Linguistics: {EMNLP} 2023},
  pages     = {5602--5621},
  year      = {2023},
  url       = {https://aclanthology.org/2023.findings-emnlp.372}
}

@inproceedings{wang-etal-2024-bridging,
  title     = {Bridging the Novice-Expert Gap via Models of Decision-Making: A Case Study on Remediating Math Mistakes},
  author    = {Wang, Rose E. and Zhang, Qingyang and Robinson, Carly and Loeb, Susanna and Demszky, Dorottya},
  booktitle = {Proceedings of the 2024 Conference of the North American Chapter of the Association for Computational Linguistics: Human Language Technologies},
  pages     = {2174--2199},
  year      = {2024},
  url       = {https://aclanthology.org/2024.naacl-long.120}
}

@inproceedings{dekmak-etal-2025-tutormind,
  title     = {{TutorMind} at {BEA} 2025 Shared Task: Leveraging Fine-Tuned {LLMs} and Data Augmentation for Mistake Identification},
  author    = {Dekmak, Fatima and Khairallah, Christian and Antoun, Wissam},
  booktitle = {Proceedings of the 20th Workshop on Innovative Use of {NLP} for Building Educational Applications},
  year      = {2025},
  url       = {https://aclanthology.org/2025.bea-1.96}
}

@inproceedings{wei-etal-2022-chain,
  title     = {Chain-of-Thought Prompting Elicits Reasoning in Large Language Models},
  author    = {Wei, Jason and Wang, Xuezhi and Schuurmans, Dale and Bosma, Maarten and Ichter, Brian and Xia, Fei and Chi, Ed and Le, Quoc and Zhou, Denny},
  booktitle = {Advances in Neural Information Processing Systems},
  volume    = {35},
  pages     = {24824--24837},
  year      = {2022},
  url       = {https://proceedings.neurips.cc/paper_files/paper/2022/file/9d5609613524ecf4f15af0f7b31abca4-Paper-Conference.pdf}
}

@misc{jurenka-etal-2024-learnlm,
  title  = {Towards Responsible Development of Generative {AI} for Education: An Evaluation-Driven Approach},
  author = {Jurenka, Irina and Kunesch, Markus and others},
  year   = {2024},
  url    = {https://arxiv.org/abs/2407.12687}
}

@misc{koutcheme-etal-2024-opensource,
  title  = {Open Source Language Models Can Provide Feedback: Evaluating {LLMs}' Ability to Help Students Using {GPT}-4-As-A-Judge},
  author = {Koutcheme, Charles and Dainese, Nicola and Sarsa, Sami and Hellas, Arto and Leinonen, Juho and Denny, Paul},
  year   = {2024},
  url    = {https://dl.acm.org/doi/10.1145/3649217.3653612}
}

@inproceedings{feng-etal-2021-survey,
  title     = {A Survey of Data Augmentation Approaches for {NLP}},
  author    = {Feng, Steven Y. and Gangal, Varun and Wei, Jason and Chandar, Sarath and Vosoughi, Soroush and Mitamura, Teruko and Hovy, Eduard},
  booktitle = {Findings of the Association for Computational Linguistics: {ACL-IJCNLP} 2021},
  pages     = {968--988},
  year      = {2021},
  url       = {https://aclanthology.org/2021.findings-acl.84}
}

@inproceedings{gombert-etal-2025-tba,
  title     = {{TBA} at {BEA} 2025 Shared Task: Transfer-Learning from {DARE-TIES} Merged Models for the Pedagogical Ability Assessment of {LLM}-Powered Math Tutors},
  author    = {Gombert, Sebastian and Zehner, Fabian and Drachsler, Hendrik},
  booktitle = {Proceedings of the 20th Workshop on Innovative Use of {NLP} for Building Educational Applications},
  pages     = {1173--1179},
  year      = {2025},
  url       = {https://aclanthology.org/2025.bea-1.92}
}

@inproceedings{saha-etal-2025-nlip,
  title     = {{NLIP} at {BEA} 2025 Shared Task: Evaluation of Pedagogical Ability of {AI} Tutors},
  author    = {Saha, Trishita and Ganguli, Shrenik and Desarkar, Maunendra Sankar},
  booktitle = {Proceedings of the 20th Workshop on Innovative Use of {NLP} for Building Educational Applications},
  pages     = {1242--1253},
  year      = {2025},
  url       = {https://aclanthology.org/2025.bea-1.99}
}

@inproceedings{fan-etal-2025-bjtu,
  title     = {{BJTU} at {BEA} 2025 Shared Task: Task-Aware Prompt Tuning and Data Augmentation for Evaluating {AI} Math Tutors},
  author    = {Fan, Yuming and Tan, Chuangchuang and Song, Wenyu},
  booktitle = {Proceedings of the 20th Workshop on Innovative Use of {NLP} for Building Educational Applications},
  year      = {2025},
  url       = {https://aclanthology.org/2025.bea-1.82/}
}

@inproceedings{hikal-etal-2025-msa,
  title     = {{MSA} at {BEA} 2025 Shared Task: Disagreement-Aware Instruction Tuning for Multi-Dimensional Evaluation of {LLMs} as Math Tutors},
  author    = {Hikal, Baraa and Basem, Mohamed and Oshallah, Islam Abdulhakeem and Hamdi, Ali},
  booktitle = {Proceedings of the 20th Workshop on Innovative Use of {NLP} for Building Educational Applications},
  year      = {2025},
  url       = {https://aclanthology.org/2025.bea-1.95/}
}

@inproceedings{roh-bang-2025-beajh,
  title     = {bea-jh at {BEA} 2025 Shared Task: Evaluating {AI}-powered Tutors through Pedagogically-Informed Reasoning},
  author    = {Roh, Jihyeon and Bang, Jinhyun},
  booktitle = {Proceedings of the 20th Workshop on Innovative Use of {NLP} for Building Educational Applications},
  year      = {2025},
  url       = {https://aclanthology.org/2025.bea-1.93}
}

@inproceedings{an-etal-2025-blcu,
  title     = {{BLCU}-{ICALL} at {BEA} 2025 Shared Task: Multi-Strategy Evaluation of {AI} Tutors},
  author    = {An, Jiyuan and Fu, Xiang and Liu, Bo and Zong, Xuquan and Kong, Cunliang and Liu, Shuliang and Wang, Shuo and Liu, Zhenghao and Yang, Liner and Fan, Hanghang and Yang, Erhong},
  booktitle = {Proceedings of the 20th Workshop on Innovative Use of {NLP} for Building Educational Applications},
  year      = {2025},
  url       = {https://aclanthology.org/2025.bea-1.84},
  doi       = {10.18653/v1/2025.bea-1.84},
  pages     = {1084--1097}
}

@inproceedings{park-etal-2025-k,
  title     = {K-{NLP}ers at {BEA} 2025 Shared Task: Evaluating the Quality of {AI} Tutor Responses with {GPT}-4.1},
  author    = {Park, Geon and Song, Jiwoo and Choi, Gihyeon and Sun, Juoh and Kim, Harksoo},
  booktitle = {Proceedings of the 20th Workshop on Innovative Use of {NLP} for Building Educational Applications},
  year      = {2025},
  url       = {https://aclanthology.org/2025.bea-1.90},
  doi       = {10.18653/v1/2025.bea-1.90},
  pages     = {1145--1163}
}

@inproceedings{hu-etal-2022-lora,
  title     = {{LoRA}: Low-Rank Adaptation of Large Language Models},
  author    = {Hu, Edward J. and Shen, Yelong and Wallis, Phillip and
               Allen-Zhu, Zeyuan and Li, Yuanzhi and Wang, Shean and
               Wang, Lu and Chen, Weizhu},
  booktitle = {International Conference on Learning Representations},
  year      = {2022},
  url       = {https://openreview.net/forum?id=nZeVKeeFYf9}
}

@inproceedings{loshchilov-hutter-2019-adamw,
  title     = {Decoupled Weight Decay Regularization},
  author    = {Loshchilov, Ilya and Hutter, Frank},
  booktitle = {International Conference on Learning Representations},
  year      = {2019},
  url       = {https://openreview.net/forum?id=Bkg6RiCqY7}
}

@misc{han-etal-2023-unsloth,
  author = {Han, Daniel and Han, Michael and {Unsloth team}},
  title  = {Unsloth},
  year   = {2023},
  url    = {https://github.com/unslothai/unsloth}
}

@misc{dubey-etal-2024-llama3,
  title         = {The {Llama} 3 Herd of Models},
  author        = {Dubey, Abhimanyu and others},
  year          = {2024},
  eprint        = {2407.21783},
  archivePrefix = {arXiv},
  url           = {https://ai.meta.com/research/publications/the-llama-3-herd-of-models/}
}

@misc{jiang-etal-2023-mistral,
  title         = {Mistral {7B}},
  author        = {Jiang, Albert Q. and Sablayrolles, Alexandre and
                   Mensch, Arthur and Bamford, Chris and Chaplot,
                   Devendra Singh and de las Casas, Diego and
                   Bressand, Florian and Lengyel, Gianna and
                   Lample, Guillaume and Saulnier, Lucile and
                   Lavaud, L{\'e}lio Renard and Lachaux, Marie-Anne
                   and Stock, Pierre and Scao, Teven Le and
                   Lavril, Thibaut and Wang, Thomas and
                   Lacroix, Timoth{\'e}e and Sayed, William El},
  year          = {2023},
  eprint        = {2310.06825},
  archivePrefix = {arXiv},
  url           = {https://arxiv.org/abs/2310.06825}
}

@misc{qwen-team-2025-qwen3,
  title         = {Qwen3 Technical Report},
  author        = {{Qwen Team}},
  year          = {2025},
  eprint        = {2505.09388},
  archivePrefix = {arXiv},
  url           = {https://arxiv.org/abs/2505.09388}
}

@misc{gemma-team-2025-gemma3,
  title         = {Gemma 3 Technical Report},
  author        = {{Gemma Team}},
  year          = {2025},
  eprint        = {2503.19786},
  archivePrefix = {arXiv},
  url           = {https://arxiv.org/abs/2503.19786}
}

@inproceedings{strubell-etal-2019-energy,
    title = "Energy and Policy Considerations for Deep Learning in {NLP}",
    author = "Strubell, Emma  and
      Ganesh, Ananya  and
      McCallum, Andrew",
    booktitle = "Proceedings of the 57th Annual Meeting of the Association for Computational Linguistics",
    month = jul,
    year = "2019",
    address = "Florence, Italy",
    publisher = "Association for Computational Linguistics",
    url = "https://aclanthology.org/P19-1355",
    doi = "10.18653/v1/P19-1355",
    pages = "3645--3650",
}

@misc{courty2024codecarbon,
  author       = {Benoit Courty and
                  Victor Schmidt and
                  Sasha Luccioni and
                  Goyal-Kamal and
                  Boris Feld and
                  J{\'e}r{\'e}my Lecourt and
                  Amine Saboni and
                  Mathilde L{\'e}val and
                  Luis Blanche and
                  Franklin Zhao and
                  Aditya Joshi},
  title        = {{CodeCarbon: Estimate and Track Carbon Emissions
                   from Machine Learning Computing}},
  version      = {3.2.6},
  year         = {2024},
  publisher    = {Zenodo},
  doi          = {10.5281/zenodo.11171501},
  url          = {https://doi.org/10.5281/zenodo.11171501}
}

@misc{deepseek-ai-2025-r1,
  title        = {DeepSeek-R1},
  author       = {{DeepSeek-AI}},
  year         = {2025},
  howpublished = {\url{https://huggingface.co/deepseek-ai/DeepSeek-R1}},
  note         = {Hugging Face model repository}
}

\clearpage
\appendix
\onecolumn
\section{Prompts}

\begin{table}[h]
\renewcommand{\arraystretch}{1.5}
\small
\begin{tabularx}{\textwidth}{>{\bfseries}p{1.0cm} X}
\toprule
\textbf{Task} & \textbf{System Prompt} \\
\midrule

MI & Classify the tutor's response to the student's answer based on 
whether the tutor has identified a mistake. Use the following labels: 
\textit{Yes} means the mistake is clearly identified; \textit{No} means 
the tutor does not recognize the mistake; \textit{To some extent} means 
the tutor suggests a mistake but is unsure. Respond strictly with exactly 
one of the following:\newline
\texttt{Evaluation: Yes}\newline
\texttt{Evaluation: No}\newline
\texttt{Evaluation: To some extent} \\

\midrule

ML & Classify the tutor's response to the student's answer based on 
whether the tutor has correctly located where the student's mistake 
occurred. Use the following labels: \textit{Yes} means the mistake 
location is clearly and correctly identified; \textit{No} means the tutor 
does not locate or pinpoint the mistake at all; \textit{To some extent} 
means the tutor partially or vaguely locates the mistake. Respond strictly 
with exactly one of the following:\newline
\texttt{Evaluation: Yes}\newline
\texttt{Evaluation: No}\newline
\texttt{Evaluation: To some extent} \\

\midrule

PG & Classify the tutor's response to the student's answer based on 
whether the tutor provides useful guidance to help the student correct 
their mistake. Use the following labels: \textit{Yes} means clear, useful, 
and relevant guidance is provided; \textit{No} means no meaningful 
guidance is provided; \textit{To some extent} means some guidance is 
provided but it is incomplete, vague, or only partially helpful. Respond 
strictly with exactly one of the following:\newline
\texttt{Evaluation: Yes}\newline
\texttt{Evaluation: No}\newline
\texttt{Evaluation: To some extent} \\

\midrule

Act & Classify the tutor's response to the student's answer based on 
whether the tutor's response is actionable---i.e., it gives the student 
something concrete and clear to do next. Use the following labels: 
\textit{Yes} means the response is clearly actionable with a concrete next 
step; \textit{No} means the response is not actionable and gives the 
student nothing to act on; \textit{To some extent} means the response is 
partially actionable but lacks clarity or completeness. Respond strictly 
with exactly one of the following:\newline
\texttt{Evaluation: Yes}\newline
\texttt{Evaluation: No}\newline
\texttt{Evaluation: To some extent} \\

\midrule

MT & You are an expert educational evaluator. Given a student-tutor math 
dialogue and a tutor response, evaluate the tutor response across four 
pedagogical dimensions. Evaluate each dimension independently:\newline
\texttt{1.\ Mistake\_Identification:} Has the tutor identified the 
student's mistake?\newline
\texttt{2.\ Mistake\_Location:} Has the tutor correctly located where the 
mistake occurred?\newline
\texttt{3.\ Providing\_Guidance:} Has the tutor provided useful guidance 
to correct the mistake?\newline
\texttt{4.\ Actionability:} Is the tutor's response actionable---does it 
give the student a concrete next step?\newline
For each dimension, use one of: \textit{Yes} / \textit{No} / \textit{To 
some extent}. Respond strictly in the following format:\newline
\texttt{Mistake\_Identification: Yes}\newline
\texttt{Mistake\_Location: No}\newline
\texttt{Providing\_Guidance: To some extent}\newline
\texttt{Actionability: Yes}\newline
Pick exactly one value per dimension. \\

\bottomrule
\end{tabularx}
\caption{Standard classification prompts used across all models in zero-shot and LoRA fine-tuning experiments. Single-task prompts produce one line output for the specific task; the multitask prompt produces four dimension-specific lines.}
\label{tab:prompts-standard}
\end{table}

\begin{table}[h]
\renewcommand{\arraystretch}{1.5}
\small
\begin{tabularx}{\textwidth}{>{\bfseries}p{1.0cm} X}
\toprule
\textbf{Task} & \textbf{System Prompt} \\
\midrule

MI & Classify the tutor's response to the student's answer based on 
whether the tutor has identified a mistake. Think step-by-step: first 
analyze the student's response to understand what mistake was made, then 
examine the tutor's response to determine if and how clearly the mistake 
is identified. Use the following labels: \textit{Yes} means the mistake 
is clearly identified; \textit{No} means the tutor does not recognize the 
mistake; \textit{To some extent} means the tutor suggests a mistake but 
is unsure. After your reasoning, respond strictly with exactly one of the 
following:\newline
\texttt{Evaluation: Yes}\newline
\texttt{Evaluation: No}\newline
\texttt{Evaluation: To some extent} \\

\midrule

ML & Classify the tutor's response to the student's answer based on 
whether the tutor has correctly located where the student's mistake 
occurred. Think step-by-step: first identify where in the student's 
solution the mistake is, then examine whether the tutor's response 
pinpoints this exact location. Use the following labels: \textit{Yes} 
means the mistake location is clearly and correctly identified; 
\textit{No} means the tutor does not locate or pinpoint the mistake at 
all; \textit{To some extent} means the tutor partially or vaguely locates 
the mistake. After your reasoning, respond strictly with exactly one of 
the following:\newline
\texttt{Evaluation: Yes}\newline
\texttt{Evaluation: No}\newline
\texttt{Evaluation: To some extent} \\

\midrule

PG & Classify the tutor's response to the student's answer based on 
whether the tutor provides useful guidance to help the student correct 
their mistake. Think step-by-step: first understand what the student 
needs to correct their error, then evaluate whether the tutor's response 
provides relevant and helpful guidance toward that correction. Use the 
following labels: \textit{Yes} means clear, useful, and relevant guidance 
is provided; \textit{No} means no meaningful guidance is provided; 
\textit{To some extent} means some guidance is provided but it is 
incomplete, vague, or only partially helpful. After your reasoning, 
respond strictly with exactly one of the following:\newline
\texttt{Evaluation: Yes}\newline
\texttt{Evaluation: No}\newline
\texttt{Evaluation: To some extent} \\

\midrule

Act & Classify the tutor's response to the student's answer based on 
whether the tutor's response is actionable---i.e., it gives the student 
something concrete and clear to do next. Think step-by-step: consider 
what a student would actually do after reading this tutor response---is 
there a clear next action? Is it specific enough to act on? Use the 
following labels: \textit{Yes} means the response is clearly actionable 
with a concrete next step; \textit{No} means the response is not 
actionable and gives the student nothing to act on; \textit{To some 
extent} means the response is partially actionable but lacks clarity or 
completeness. After your reasoning, respond strictly with exactly one of 
the following:\newline
\texttt{Evaluation: Yes}\newline
\texttt{Evaluation: No}\newline
\texttt{Evaluation: To some extent} \\

\midrule

MT & You are an expert educational evaluator. Given a student-tutor math 
dialogue and a tutor response, evaluate the tutor response across four 
pedagogical dimensions. Think step-by-step for each dimension: carefully 
analyze the dialogue, identify the student's mistake, and then evaluate 
each dimension independently before giving your final answer.\newline
\texttt{1.\ Mistake\_Identification:} Has the tutor identified the 
student's mistake?\newline
\texttt{2.\ Mistake\_Location:} Has the tutor correctly located where the 
mistake occurred?\newline
\texttt{3.\ Providing\_Guidance:} Has the tutor provided useful guidance 
to correct the mistake?\newline
\texttt{4.\ Actionability:} Is the tutor's response actionable---does it 
give the student a concrete next step?\newline
For each dimension, use one of: \textit{Yes} / \textit{No} / \textit{To 
some extent}. After your reasoning, respond strictly in the following 
format:\newline
\texttt{Mistake\_Identification: Yes}\newline
\texttt{Mistake\_Location: No}\newline
\texttt{Providing\_Guidance: To some extent}\newline
\texttt{Actionability: Yes}\newline
Pick exactly one value per dimension. \\

\bottomrule
\end{tabularx}
\caption{Chain-of-thought classification prompts used for 
Qwen3-14B in zero-shot and LoRA with thinking ON. Identical in structure to Table~\ref{tab:prompts-standard} 
but augmented with explicit step-by-step reasoning instructions before 
the final label prediction.}
\label{tab:prompts-cot}
\end{table}

\clearpage
\small
\renewcommand{\arraystretch}{1.5}
\begin{longtable}{>{\bfseries}p{1.0cm} >{\bfseries}p{2.2cm} p{\dimexpr\textwidth-3.2cm-6\tabcolsep}}

\toprule
\textbf{Task} & \textbf{Minority Class} & \textbf{Generation Prompt} \\
\midrule
\endfirsthead

\toprule
\textbf{Task} & \textbf{Minority Class} & \textbf{Generation Prompt} \\
\midrule
\endhead

MI & No & You are an expert math tutor generating training data for an AI 
evaluation system. Given the following student-tutor math conversation, 
write a single tutor response that does NOT identify the student's mistake 
at all. The tutor should either ignore the mistake, proceed as if the 
student is correct, or simply provide the next step without any 
acknowledgment of an error. The response must be exactly ONE sentence, 
natural and realistic---something an actual tutor might say. Write only 
the single-sentence tutor response, nothing else. \\
\midrule

MI & To some extent & You are an expert math tutor generating training 
data for an AI evaluation system. Given the following student-tutor math 
conversation, write a single tutor response that PARTIALLY or VAGUELY 
suggests the student may have made a mistake, but does not clearly or 
explicitly identify what the mistake is. The tutor should sound uncertain, 
exploratory, or cautious. The response must be exactly ONE sentence, 
natural and realistic---something an actual tutor might say. Write only 
the single-sentence tutor response, nothing else. \\
\midrule

ML & No & You are an expert math tutor generating training data for an AI 
evaluation system. Given the following student-tutor math conversation, 
write a single tutor response that does NOT locate or pinpoint where the 
student's mistake occurred. The tutor may acknowledge something is off but 
gives no indication of where in the solution the error is. The response 
must be exactly ONE sentence, natural and realistic---something an actual 
tutor might say. Write only the single-sentence tutor response, nothing 
else. \\
\midrule

ML & To some extent & You are an expert math tutor generating training 
data for an AI evaluation system. Given the following student-tutor math 
conversation, write a single tutor response that PARTIALLY locates where 
the student's mistake is, but is vague, imprecise, or only hints at the 
location without clearly identifying it. The response must be exactly ONE 
sentence, natural and realistic---something an actual tutor might say. 
Write only the single-sentence tutor response, nothing else. \\
\midrule

PG & No & You are an expert math tutor generating training data for an AI 
evaluation system. Given the following student-tutor math conversation, 
write a single tutor response that provides NO useful guidance to help the 
student correct their mistake. The tutor might point out an error but 
gives the student nothing helpful to act on, or simply restates the 
problem without direction. The response must be exactly ONE sentence, 
natural and realistic---something an actual tutor might say. Write only 
the single-sentence tutor response, nothing else. \\
\midrule

PG & To some extent & You are an expert math tutor generating training 
data for an AI evaluation system. Given the following student-tutor math 
conversation, write a single tutor response that provides SOME guidance 
but is incomplete, too vague, or only partially helpful. The student would 
have some direction but not enough to fully correct their mistake. The 
response must be exactly ONE sentence, natural and realistic---something 
an actual tutor might say. Write only the single-sentence tutor response, 
nothing else. \\
\midrule

Act & No & You are an expert math tutor generating training data for an 
AI evaluation system. Given the following student-tutor math conversation, 
write a single tutor response that is NOT actionable---it gives the 
student nothing concrete to do next. The response might be motivational 
or general but lacks any specific next step. The response must be exactly 
ONE sentence, natural and realistic---something an actual tutor might say. 
Write only the single-sentence tutor response, nothing else. \\
\midrule

Act & To some extent & You are an expert math tutor generating training 
data for an AI evaluation system. Given the following student-tutor math 
conversation, write a single tutor response that is PARTIALLY 
actionable---it gives the student some direction but the next step is 
unclear, incomplete, or ambiguous. The response must be exactly ONE 
sentence, natural and realistic---something an actual tutor might say. 
Write only the single-sentence tutor response, nothing else. \\
\midrule

MT & No (all dims) & You are an expert math tutor generating training data 
for an AI evaluation system. Given the following student-tutor math 
conversation, write a single tutor response that scores \textit{No} on 
ALL four pedagogical dimensions: it does not identify the mistake, does 
not locate it, provides no guidance, and is not actionable. The response 
must be exactly ONE sentence, natural and realistic---something an actual 
tutor might say. Write only the single-sentence tutor response, nothing 
else. \\
\midrule

MT & To some extent (all dims) & You are an expert math tutor generating 
training data for an AI evaluation system. Given the following 
student-tutor math conversation, write a single tutor response that scores 
\textit{To some extent} on ALL four pedagogical dimensions: it vaguely 
suggests a mistake, partially locates it, gives incomplete guidance, and 
is only partially actionable. The response must be exactly ONE sentence, 
natural and realistic---something an actual tutor might say. Write only 
the single-sentence tutor response, nothing else. \\

\bottomrule
\caption[]{Augmentation generation prompts used by Qwen3-14B (thinking ON) 
to synthesize minority-class examples (\textit{No} and \textit{To some 
extent}). Responses are constrained to one sentence 
to match real tutor response style.}
\label{tab:prompts-augmentation}

\end{longtable}

\begin{table}[h]
\renewcommand{\arraystretch}{1.5}
\small
\begin{tabularx}{\textwidth}{>{\bfseries}p{1.0cm} X}
\toprule
\textbf{Task} & \textbf{Verification Prompt} \\
\midrule

MI & You are an expert educational evaluator. Given a student-tutor math conversation and a generated tutor response, classify whether the tutor response identifies the student's mistake. Labels: \textit{Yes}: The tutor clearly identifies or recognizes the student's mistake. \textit{No}: The tutor does not recognize or acknowledge any mistake. \textit{To some 
extent}: The tutor vaguely, cautiously, or partially suggests there may be a mistake but does not clearly identify it. Think carefully, but output only one line using exactly one of the following:\newline
\texttt{Evaluation: Yes}\newline
\texttt{Evaluation: No}\newline
\texttt{Evaluation: To some extent} \\
\midrule

ML & You are an expert educational evaluator. Given a student-tutor math conversation and a generated tutor response, classify whether the tutor response locates where the student's mistake occurred.Labels: \textit{Yes}: The tutor clearly and correctly pinpoints where the mistake occurred. \textit{No}: The tutor does not locate or pinpoint the mistake at all. \textit{To some 
extent}: The tutor gives only a vague, broad, partial, or imprecise indication of where the mistake may be. Think carefully, but output only one line using exactly one of the following:\newline
\texttt{Evaluation: Yes}\newline
\texttt{Evaluation: No}\newline
\texttt{Evaluation: To some extent} \\
\midrule

PG & You are an expert educational evaluator. Given a student-tutor math conversation and a generated tutor response, classify whether the tutor response provides useful guidance to help the student correct the mistake. Labels: \textit{Yes}: The tutor provides clear, useful, and relevant guidance sufficient to help the student correct the mistake. \textit{No}: The tutor provides no meaningful guidance, hint, strategy, correction, or useful direction. \textit{To some 
extent}: The tutor provides some guidance, but it is incomplete, vague, under-specified, or only partially helpful. Think carefully, but output only one line using exactly one of the following:\newline
\texttt{Evaluation: Yes}\newline
\texttt{Evaluation: No}\newline
\texttt{Evaluation: To some extent} \\
\midrule

Act & You are an expert educational evaluator. Given a student-tutor math conversation and a generated tutor response, classify whether the tutor response gives the student a concrete next action. Labels: \textit{Yes}: The tutor gives a clear and concrete next step the student can act on. \textit{No}: The tutor gives no concrete action or next step.\textit{To some 
extent}: The tutor gives a vague, incomplete, or partially actionable next step, but it is not specific enough to be fully actionable. Think carefully, but output only one line using exactly one of the following:\newline
\texttt{Evaluation: Yes}\newline
\texttt{Evaluation: No}\newline
\texttt{Evaluation: To some extent} \\

\bottomrule
\end{tabularx}
\caption{Self-verification prompts used by Qwen3-14B to filter 
synthetic examples generated during augmentation. 
The placeholder \texttt{\{label\}} is replaced at runtime with the 
intended minority class label. For multitask examples, all four prompts 
run independently.}
\label{tab:prompts-verification}
\end{table}

\section{Lenient F1 Results}
\label{sec:lenient-results}

Tables~\ref{tab:lenient-zs-single} through~\ref{tab:lenient-think-on-mt} report 
lenient macro-averaged F1 scores across all experimental conditions. G3-12B and 
G3-27B denote Gemma3-12B and Gemma3-27B (8-bit), respectively. Best value per 
dimension is bolded in each table.

\begin{table}[h]
\centering
\footnotesize
\setlength{\tabcolsep}{4pt}
\renewcommand{\arraystretch}{1.1}
\begin{tabular}{lrrrr}
\toprule
\textbf{Model} & \textbf{MI} & \textbf{ML} & \textbf{PG} & \textbf{Act} \\
\midrule
LLaMA-3.1-8B & 0.7662 & 0.6748 & 0.6631 & 0.6265 \\
Mistral-7B   & 0.7621 & 0.5393 & 0.5188 & 0.4945 \\
Qwen3-14B    & 0.7747 & \textbf{0.6813} & 0.6332 & 0.6204 \\
G3-12B       & \textbf{0.7761} & 0.6795 & \textbf{0.7131} & 0.6258 \\
G3-27B       & 0.7444 & 0.6645 & 0.6762 & \textbf{0.6644} \\
\bottomrule
\end{tabular}
\caption{Zero-shot | No-Aug | Think OFF | Single Tasks | Lenient F1.}
\label{tab:lenient-zs-single}
\end{table}

\begin{table}[h]
\centering
\footnotesize
\setlength{\tabcolsep}{4pt}
\renewcommand{\arraystretch}{1.1}
\begin{tabular}{lrrrr}
\toprule
\textbf{Model} & \textbf{MI} & \textbf{ML} & \textbf{PG} & \textbf{Act} \\
\midrule
LLaMA-3.1-8B & 0.6962 & 0.5213 & 0.5907 & 0.5067 \\
Mistral-7B   & 0.5186 & 0.4821 & 0.4687 & 0.4109 \\
Qwen3-14B    & 0.6990 & 0.5777 & 0.7178 & 0.5766 \\
G3-12B       & 0.7280 & 0.6031 & 0.7234 & 0.6770 \\
G3-27B       & \textbf{0.7587} & \textbf{0.6630} & \textbf{0.7257} & \textbf{0.6764} \\
\bottomrule
\end{tabular}
\caption{Zero-shot | No-Aug | Think OFF | Multitask | Lenient F1.}
\label{tab:lenient-zs-mt}
\end{table}

\begin{table}[h]
\centering
\footnotesize
\setlength{\tabcolsep}{4pt}
\renewcommand{\arraystretch}{1.1}
\begin{tabular}{lrr}
\toprule
\textbf{Dimension} & \textbf{Think OFF} & \textbf{Think ON} \\
\midrule
MI  & \textbf{0.7747} & 0.6820 \\
ML  & \textbf{0.6813} & 0.6750 \\
PG  & 0.6332 & \textbf{0.6517} \\
Act & \textbf{0.6204} & 0.5792 \\
\bottomrule
\end{tabular}
\caption{Zero-shot | No-Aug | Qwen3-14B | Think OFF vs ON | Single Tasks | 
Lenient F1. No MT data available for Think ON.}
\label{tab:lenient-zs-think}
\end{table}

\begin{table}[h]
\centering
\footnotesize
\setlength{\tabcolsep}{4pt}
\renewcommand{\arraystretch}{1.1}
\begin{tabular}{lrrrr}
\toprule
\textbf{Model} & \textbf{MI} & \textbf{ML} & \textbf{PG} & \textbf{Act} \\
\midrule
LLaMA-3.1-8B & 0.8685 & 0.7232 & 0.7307 & 0.7809 \\
Mistral-7B   & 0.7056 & 0.5696 & 0.5246 & 0.4702 \\
Qwen3-14B    & 0.8886 & 0.7416 & 0.7400 & 0.8151 \\
G3-12B       & \textbf{0.8948} & \textbf{0.7605} & \textbf{0.7890} & \textbf{0.8828} \\
G3-27B       & 0.4883 & 0.3379 & 0.7443 & 0.8692 \\
\bottomrule
\end{tabular}
\caption{LoRA | No-Aug | Think OFF | Single Tasks | Lenient F1.}
\label{tab:lenient-lora-single}
\end{table}

\begin{table}[h]
\centering
\footnotesize
\setlength{\tabcolsep}{4pt}
\renewcommand{\arraystretch}{1.1}
\begin{tabular}{lrrrr}
\toprule
\textbf{Model} & \textbf{MI} & \textbf{ML} & \textbf{PG} & \textbf{Act} \\
\midrule
LLaMA-3.1-8B & 0.8455 & 0.6690 & 0.7229 & 0.7767 \\
Mistral-7B   & 0.4832 & 0.4897 & 0.4679 & 0.4210 \\
Qwen3-14B    & 0.8081 & 0.6973 & 0.7307 & 0.8372 \\
G3-12B       & 0.8550 & 0.7772 & 0.7734 & 0.8736 \\
G3-27B       & \textbf{0.8751} & \textbf{0.7848} & \textbf{0.7684} & \textbf{0.8877} \\
\bottomrule
\end{tabular}
\caption{LoRA | No-Aug | Think OFF | Multitask | Lenient F1.}
\label{tab:lenient-lora-mt}
\end{table}

\begin{table}[h]
\centering
\footnotesize
\setlength{\tabcolsep}{4pt}
\renewcommand{\arraystretch}{1.1}
\begin{tabular}{lrr}
\toprule
\textbf{Dimension} & \textbf{Think OFF} & \textbf{Think ON} \\
\midrule
MI  & \textbf{0.8886} & 0.3373 \\
ML  & \textbf{0.7416} & 0.4368 \\
PG  & \textbf{0.7400} & 0.3763 \\
Act & \textbf{0.8151} & 0.4300 \\
\bottomrule
\end{tabular}
\caption{LoRA | No-Aug | Qwen3-14B | Think OFF vs ON | Single Tasks | 
Lenient F1. No MT data available for No-Aug Think ON.}
\label{tab:lenient-lora-think}
\end{table}

\begin{table*}[h]
\centering
\footnotesize
\setlength{\tabcolsep}{3pt}
\renewcommand{\arraystretch}{1.1}
\begin{tabular}{l rrrr rrrr rrrr}
\toprule
& \multicolumn{4}{c}{\textbf{No-Aug}}
& \multicolumn{4}{c}{\textbf{Qwen3-Gen}}
& \multicolumn{4}{c}{\textbf{Qwen3-Gen+Verify}} \\
\cmidrule(lr){2-5} \cmidrule(lr){6-9} \cmidrule(lr){10-13}
\textbf{Model} & MI & ML & PG & Act & MI & ML & PG & Act & MI & ML & PG & Act \\
\midrule
LLaMA-3.1-8B & 0.8685 & 0.7232 & 0.7307 & 0.7809 & 0.8206 & 0.7131 & 0.7490 & 0.7751 & 0.8297 & 0.7241 & 0.7242 & 0.7692 \\
Mistral-7B   & 0.7056 & 0.5696 & 0.5246 & 0.4702 & 0.6882 & 0.5756 & 0.6095 & 0.6018 & 0.6971 & 0.6354 & 0.5943 & 0.6952 \\
Qwen3-14B    & 0.8886 & 0.7416 & 0.7400 & 0.8151 & 0.8576 & 0.7180 & 0.7372 & 0.8137 & 0.8659 & 0.7414 & 0.7406 & 0.8390 \\
G3-12B       & \textbf{0.8948} & 0.7605 & 0.7890 & \textbf{0.8828} & 0.8586 & 0.7637 & 0.7624 & 0.8632 & 0.8766 & 0.7644 & \textbf{0.7908} & 0.8811 \\
G3-27B       & 0.4883 & 0.3379 & 0.7443 & 0.8692 & 0.8734 & 0.7633 & 0.7867 & 0.8484 & 0.8825 & \textbf{0.7846} & 0.7641 & 0.8528 \\
\bottomrule
\end{tabular}
\caption{LoRA | Think OFF | All Augmentation Strategies | Single Tasks | Lenient F1.}
\label{tab:lenient-aug-single}
\end{table*}

\begin{table*}[h]
\centering
\footnotesize
\setlength{\tabcolsep}{3pt}
\renewcommand{\arraystretch}{1.1}
\begin{tabular}{l rrrr rrrr rrrr}
\toprule
& \multicolumn{4}{c}{\textbf{No-Aug}}
& \multicolumn{4}{c}{\textbf{Qwen3-Gen}}
& \multicolumn{4}{c}{\textbf{Qwen3-Gen+Verify}} \\
\cmidrule(lr){2-5} \cmidrule(lr){6-9} \cmidrule(lr){10-13}
\textbf{Model} & MI & ML & PG & Act & MI & ML & PG & Act & MI & ML & PG & Act \\
\midrule
LLaMA-3.1-8B & 0.8455 & 0.6690 & 0.7229 & 0.7767 & 0.8507 & 0.6632 & 0.7395 & 0.7705 & 0.8397 & 0.6698 & 0.7122 & 0.7578 \\
Mistral-7B   & 0.4832 & 0.4897 & 0.4679 & 0.4210 & 0.7155 & 0.6346 & 0.7327 & 0.6706 & 0.7135 & 0.6393 & 0.7240 & 0.6732 \\
Qwen3-14B    & 0.8081 & 0.6973 & 0.7307 & 0.8372 & 0.8394 & 0.7162 & 0.7544 & 0.8455 & 0.8569 & 0.7315 & 0.7344 & 0.8381 \\
G3-12B       & 0.8550 & 0.7772 & \textbf{0.7734} & 0.8736 & 0.7793 & 0.7493 & 0.7549 & \textbf{0.8794} & 0.8328 & 0.7870 & 0.7550 & 0.8684 \\
G3-27B       & \textbf{0.8751} & 0.7848 & 0.7684 & \textbf{0.8877} & 0.8564 & 0.7806 & 0.7627 & 0.8735 & 0.8733 & \textbf{0.7907} & 0.7648 & 0.8594 \\
\bottomrule
\end{tabular}
\caption{LoRA | Think OFF | All Augmentation Strategies | Multitask | Lenient F1.}
\label{tab:lenient-aug-mt}
\end{table*}

\begin{table}[h]
\centering
\footnotesize
\setlength{\tabcolsep}{4pt}
\renewcommand{\arraystretch}{1.1}
\begin{tabular}{lrrr}
\toprule
\textbf{Dimension} & \textbf{No-Aug} & \textbf{Qwen3-Gen} & \textbf{Qwen3-Gen+Verify} \\
\midrule
MI  & 0.3373 & \textbf{0.8699} & 0.8598 \\
ML  & 0.4368 & \textbf{0.7530} & 0.7482 \\
PG  & 0.3763 & 0.7544 & \textbf{0.7581} \\
Act & 0.4300 & 0.8632 & \textbf{0.8727} \\
\bottomrule
\end{tabular}
\caption{LoRA | Qwen3-14B | Think ON | Single Tasks | Lenient F1. 
No-Aug Think ON included as baseline.}
\label{tab:lenient-think-on-single}
\end{table}

\begin{table}[h]
\centering
\footnotesize
\setlength{\tabcolsep}{4pt}
\renewcommand{\arraystretch}{1.1}
\begin{tabular}{lrr}
\toprule
\textbf{Dimension} & \textbf{Qwen3-Gen} & \textbf{Qwen3-Gen+Verify} \\
\midrule
MI  & \textbf{0.8538} & 0.8394 \\
ML  & \textbf{0.7496} & 0.7436 \\
PG  & \textbf{0.7763} & 0.7495 \\
Act & 0.8488 & \textbf{0.8564} \\
\bottomrule
\end{tabular}
\caption{LoRA | Qwen3-14B | Think ON | Multitask | Lenient F1. 
No-Aug MT not available for Think ON.}
\label{tab:lenient-think-on-mt}
\end{table}

\end{document}